\documentclass[letterpaper]{article} 
\usepackage{aaai23}  
\usepackage{times}  
\usepackage{helvet}  
\usepackage{courier}  
\usepackage[hyphens]{url}  
\usepackage{graphicx} 
\usepackage{natbib}  
\usepackage{caption} 
\usepackage{algorithm}
\usepackage{algorithmic}
\usepackage{graphicx}
\usepackage{tikz}
\usepackage{comment}
\usepackage{amsmath,amssymb} 
\usepackage{multirow}
\usepackage{multicol}
\usepackage{hhline}
\usepackage{booktabs}
\usepackage{tabularx}
\usepackage{makecell}
\usepackage{threeparttable}
\usepackage{color}
\newcommand{\ra}[1]{\renewcommand{\arraystretch}{#1}}

\usepackage{newfloat}
\usepackage{listings}
\DeclareCaptionStyle{ruled}{labelfont=normalfont,labelsep=colon,strut=off} 
\floatstyle{ruled}
\newfloat{listing}{tb}{lst}{}
\floatname{listing}{Listing}
\nocopyright
\title{Semantic-aware One-shot Face Re-enactment with Dense Correspondence Estimation}
\author{
    Yunfan Liu\textsuperscript{\rm 1}\textsuperscript{\rm 2},
    Qi Li\textsuperscript{\rm 2},
    Zhenan Sun\textsuperscript{\rm 1}\textsuperscript{\rm 2},
    Tieniu Tan\textsuperscript{\rm 1}\textsuperscript{\rm 2}
}
\affiliations{
    \textsuperscript{\rm 1}School of Artificial Intelligence, University of Chinese Academy of Sciences, Beijing, China\\
    \textsuperscript{\rm 2}CRIPAC, Institute of Automation, Chinese Academy of Sciences, Beijing, China\\
    yunfan.liu@cripac.ia.ac.cn, \{qli, znsun, tnt\}@nlpr.ia.ac.cn
}

\usepackage{bibentry}

\begin{document}

\maketitle

\begin{abstract}
One-shot face re-enactment is a challenging task due to the identity mismatch between source and driving faces.
Specifically, the suboptimally disentangled identity information of driving subjects would inevitably interfere with the re-enactment results and lead to face shape distortion. 
To solve this problem, this paper proposes to use 3D Morphable Model (3DMM) for explicit facial semantic decomposition and identity disentanglement.
Instead of using 3D coefficients alone for re-enactment control, we take the advantage of the generative ability of 3DMM to render textured face proxies.
These proxies contain abundant yet compact geometric and semantic information of human faces, which enable us to compute the face motion field between source and driving images by estimating the dense correspondence.
In this way, we could approximate re-enactment results by warping source images according to the motion field, and a Generative Adversarial Network (GAN) is adopted to further improve the visual quality of warping results.
Extensive experiments on various datasets demonstrate the advantages of the proposed method over existing start-of-the-art benchmarks in both identity preservation and re-enactment fulfillment.
\end{abstract}

\section{Introduction}
Face re-enactment aims at animating the source face with pose and expression conveyed by driving images, and meanwhile keeping its identity information unchanged.
Due to its wide range of practical applications, face re-enactment has attracted remarkable research attention in the last few years. 
With the rapid development of Generative Adversarial Networks (GANs), numerous studies have been conducted to solve the face re-enactment problem with GAN-based models~\cite{kim2018deep,thies2016face2face,wu2018reenactgan}.
Although realistic results could be obtained, these methods are either designed only for a specific subject or require a large amount of images of the source identity, which heavily limits their application in real-world scenarios.

\begin{figure}[t]
    \centering
    \includegraphics[width=0.9\linewidth]{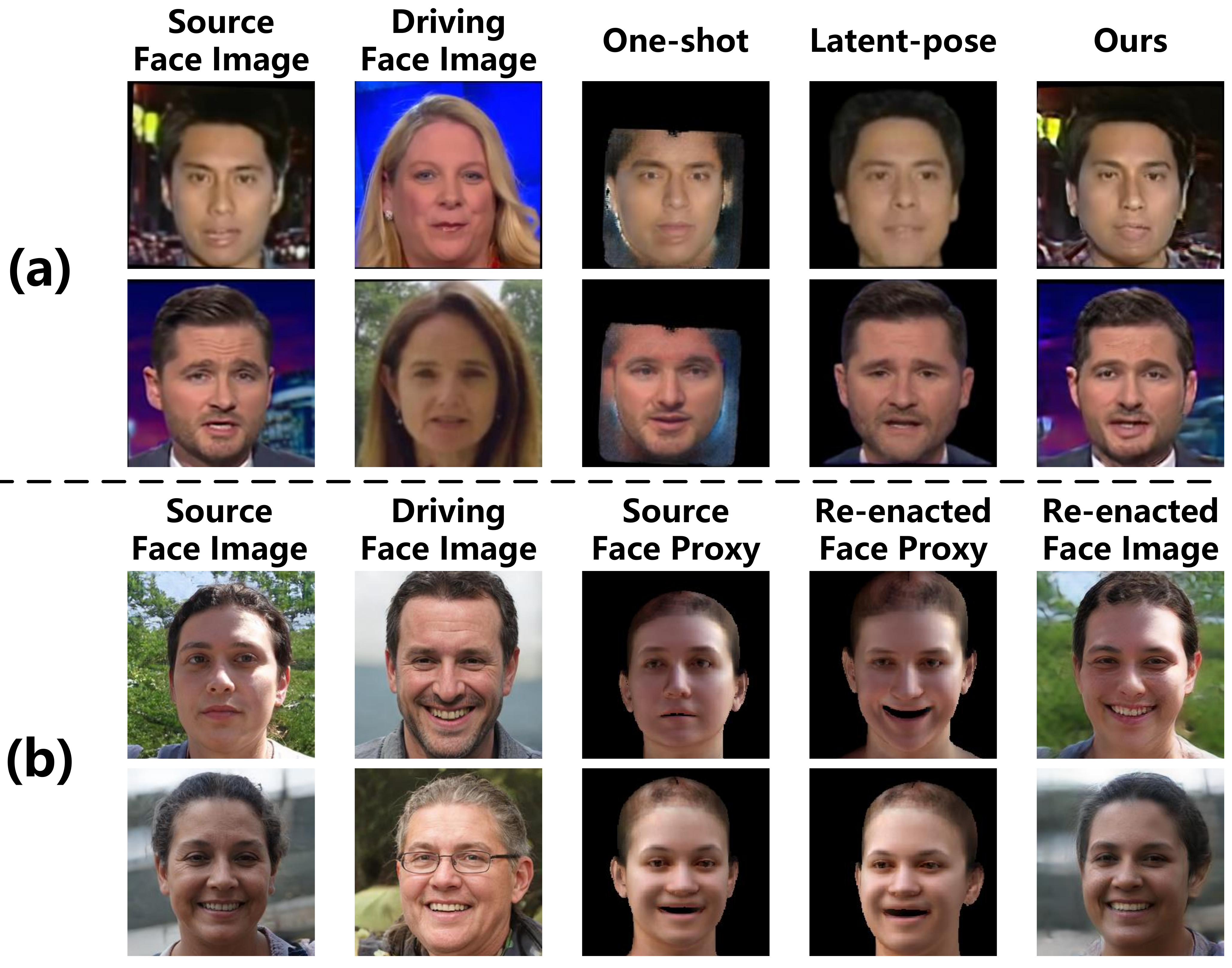}
    \caption{Illustration of (a) sample face re-enactment results with `identity leak'. Note the change of identity in results obtain from One-shot~\cite{OneShotFace2019} and Latent-pose~\cite{burkov2020neural}, and (b) visualization of the face proxy of source and re-enacted faces.}
    \label{fig:teaser}
\end{figure}

In this work, we focus on solving a more generalized problem, i.e., one-shot face re-enactment, where only one single image of the source subject is available at test time.
The key for solving this problem is to fully disentangle geometric attributes (i.e., pose and expression) in driving images from personal characteristics, otherwise it will interfere with the re-enactment process and cause shape distortion in generation results (known as the `identity leak' problem~\cite{ha2020marionette}, see Fig.~\ref{fig:teaser}(a)).

To this end, various representations for facial geometry are proposed to eliminate the influence of driving identity. \textit{Facial Action Units (FAUs)}, a coding system for classifying facial muscle movement, are used in some previous studies~\cite{pumarola2018ganimation,tripathy2020icface,tripathy2021facegan} to describe face motion.
However, the flexibility of FAUs is quite limited as they could not describe complex facial expression or head pose. 
\textit{Facial landmarks} are also widely used in existing methods~\cite{ha2020marionette,huang2020learning,nirkin2019fsgan,song2018geometry,wu2018reenactgan,zhang2020freenet,OneShotFace2019,siarohin2019first,tripathy2022single} to depict the contour and layout information of facial components with facial texture excluded.
Compared to FAUs, facial landmarks could provide a more comprehensive description of the spatial structure of human faces.
Unfortunately, they still contain personal attributes of driving subjects in terms of face shape, which may impair the performance of source identity preservation.

In this paper, we propose to use an advanced \textit{3D Morphable Model (3DMM)}, i.e., FLAME~\cite{FLAME:SiggraphAsia2017}, for facial geometry decomposition.
Unlike most existing methods which only use low-dimensional 3DMM coefficients~\cite{yin2017towards,nguyen2019hologan,tewari2020stylerig,ghosh2020gif} or plain face meshes~\cite{yao2020mesh} for face re-enactment control, we also make use of its generative ability to render \textbf{textured face proxies} (see Fig.~\ref{fig:teaser} (b)).
These representations not only contain richer spatial information of skin area and facial components, but also are less subject to the influence of high-frequency image details or background content when estimating the face motion, compared to original face images.

\begin{figure}[t]
    \centering
    \includegraphics[width=0.85\linewidth]{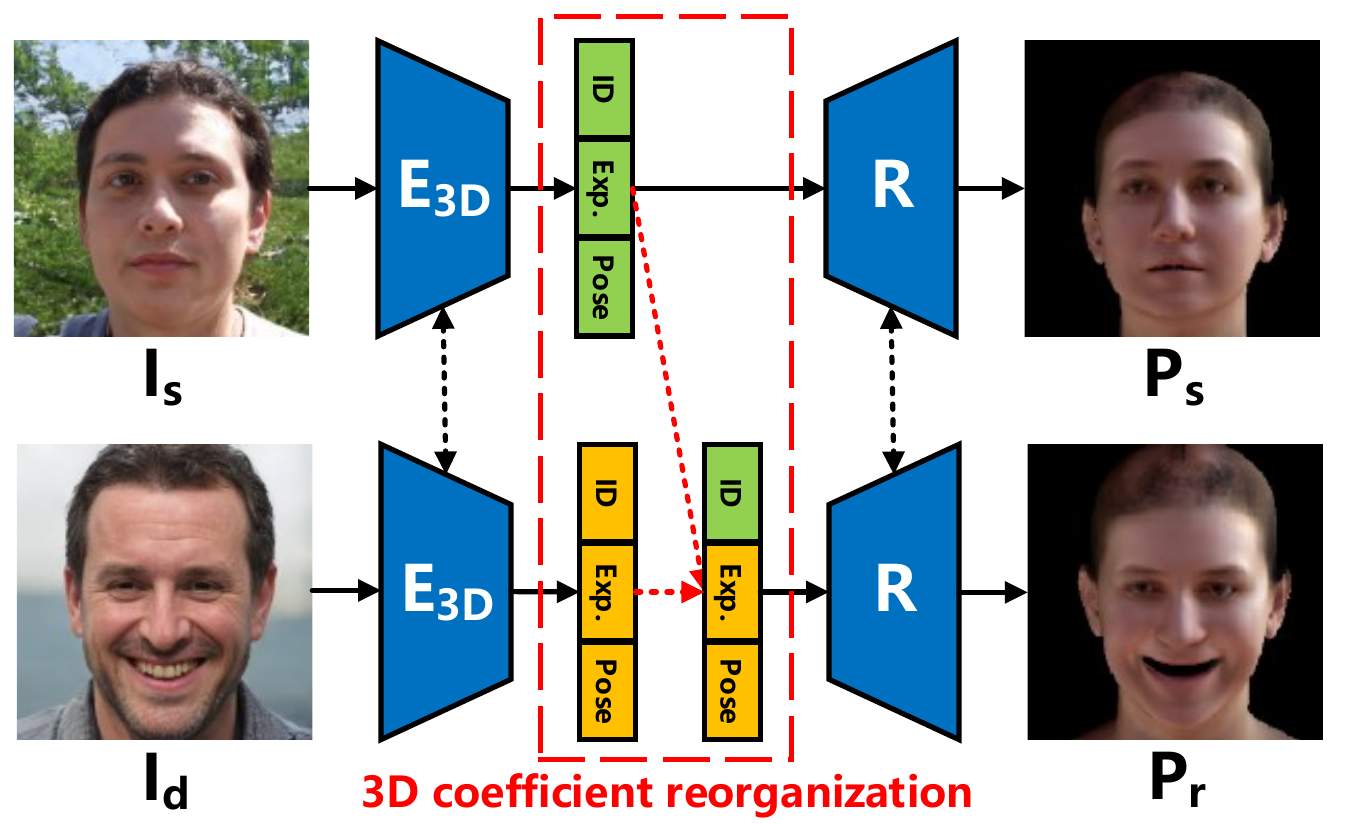}
    \caption{Illustration of the reorganization of 3D coefficients and rendering of textured face proxies. Black dashed lines with arrows denote that the connected networks (i.e., $E_{3D}$ and $R$) share the same weight.}
    \label{fig:FLAME}
\end{figure}

This inspires us to leverage the advantage of textured face proxies for face motion field estimation, which could further facilitate the prediction of re-enacted faces.
Concretely, we first generate the proxy of re-enacted face by the FLAME model based on the mixture of 3DMM parameters of source and driving faces (see Fig.~\ref{fig:FLAME}).
Then, the face motion field is computed via the proposed \textbf{D}ense \textbf{C}orrespondence \textbf{G}uided GAN (\textbf{DCG-GAN}) model via estimating the dense spatial correspondence between the proxy of source and re-enacted faces.
Afterwards, source images are warped according to the motion field, and results are fed in to a generator network to improve the visual quality.
Extensive experiment results demonstrate the effectiveness of the proposed method and its advantages over existing state-of-the-art approaches.

Contributions of this work are summarized as follows,
\begin{itemize}
    \item An advanced 3DMM model, i.e., FLAME, is adopted to disentangle the identity feature from pose and expression. We also use it to render textured face proxies to predict the geometry and semantic of re-enacted faces, which facilitate the estimation of face motion.
    
    \item A novel GAN-based framework, named DCG-GAN, is proposed to generate realistic re-enactment results. It first estimates the face motion field by computing the dense correspondence, and improve the visual quality based on the warping result.
    
    \item Extensive experiments are conducted on multiple datasets to demonstrate the advantage of DCG-GAN in solving the one-shot face re-enactment problem, compared to existing state-of-the-art methods.
\end{itemize}

\section{Method}\label{sec:method}
Given a source image $I_s$ and a driving image $I_d$, we aim to generate the re-enacted image $I_r$ with pose and expression conveyed in $I_d$ and identity in $I_s$.
As shown in Fig.~\ref{fig:teaser}, our method is generally comprised of three modules.
Firstly, a \textbf{3D-guided Face Proxy Editor} is adopted to decompose geometric semantics of both $I_s$ and $I_d$, and render the proxy of $I_s$ and $I_r$ (Section~\ref{sec:3d_face_mani_module}).
Then, the spatial correlation $\mathcal{F}$ between proxies is estimated by a \textbf{Dense Correspondence Estimator} to present the face motion, which is used to approximate $I_r$ by warping $I_s$ (Section~\ref{sec:dense_corr_esti}).
Thereafter, the warped $I_s$ and other associated spatial representations are fed into the \textbf{Facial Prior-guided Generator} to synthesize $I_r$ with high visual fidelity (Section~\ref{sec:facial_prior_guided_generator}).

\subsection{3D-guided Face Proxy Editor}\label{sec:3d_face_mani_module}
We adopt an advanced 3D parametric model, FLAME~\cite{FLAME:SiggraphAsia2017}, to explicitly decompose face semantics into various disentangled components. 
Specifically, FLAME depicts the geometry of a portrait image with coefficients of camera $\alpha$, pose $\beta$, expression $\theta$, shape $\phi$, lighting condition $\lambda$ and facial texture $\mu$.
As shown in Fig.~\ref{fig:FLAME}, an encoder network, denoted as $E_{3D}:I\rightarrow C$ ($C=\{\alpha, \beta, \theta, \phi, \lambda, \mu\}$), is proposed to regress FLAME coefficients $C$ from a given face image $I$.
Concretely, coefficients for disentangled semantics of $I_s$ and $I_d$ could be computed as $C_s=E_{3D}(I_s)$ and $C_d=E_{3D}(I_d)$, respectively.
Thus, the set of coefficients for the re-enacted face $I_r$ could be obtain by simply mixing the element in both $C_s$ and $C_d$, which could be denoted as $C_r=\{\alpha_d, \beta_d, \theta_d, \phi_s, \lambda_s, \mu_s\}$ (geometry-related coefficients obtained from $I_d$, and the rest from $I_s$).

\begin{figure*}[t]
    \centering
    \includegraphics[width=0.85\linewidth]{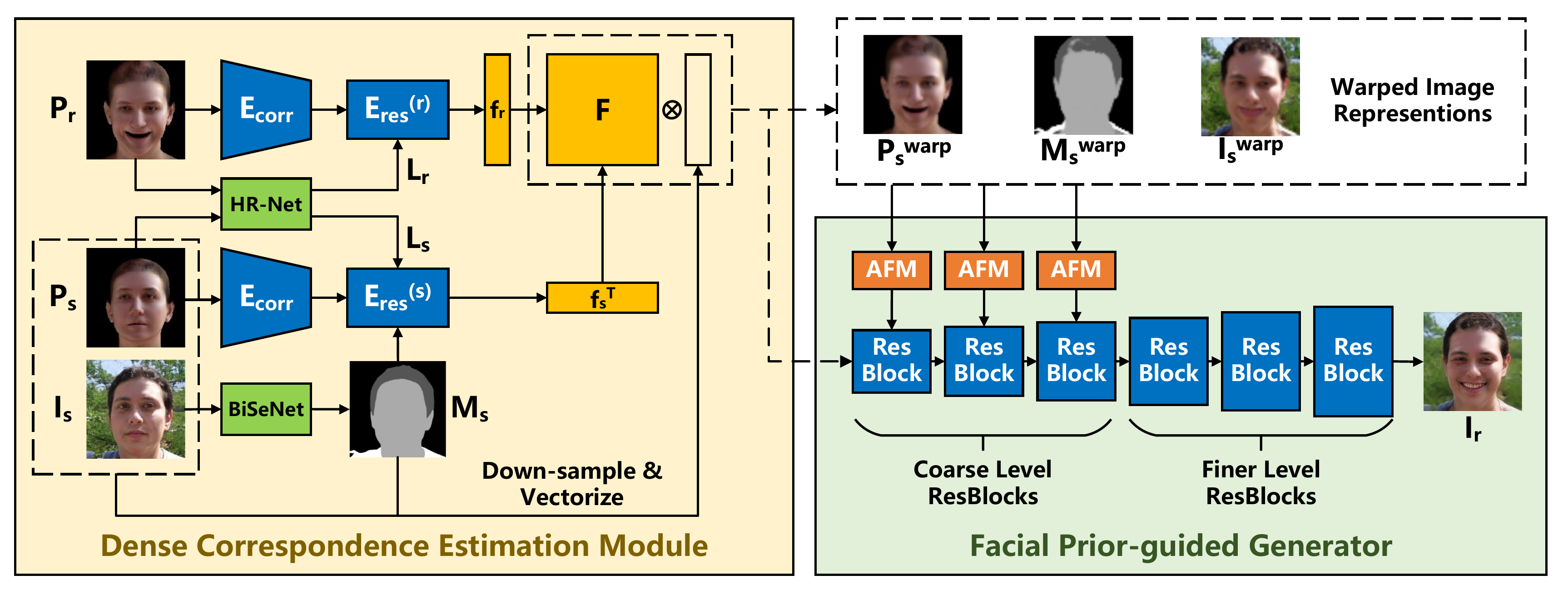}
    \caption{An overview of the proposed model DCG-GAN. The Dense Correspondence Estimation Module computes the correlation matrix based on feature vectors extracted from face proxies. Afterwards, warped spatial representations of the source face are fed into the Facial Prior-guided Generator to synthesize re-enactment results with high quality.}
    \label{fig:SSS-GAN}
\end{figure*}

Unlike most existing methods merely using low-dimensional 3DMM coefficients for re-enactment control, we also make use of the generative ability of FLAME to convey more semantic information in the spatial dimension.
To this end, given the set of parameters $C_r$, a differentiable rendering network~\cite{DECA:Siggraph2021} (denoted as $R:C\rightarrow I$) is adopted to generate the corresponding proxy image $P_r=R(C_r)$.
Notably, $P_r$ helps mitigate the identity gap as it shares the same identity with $P_s=R(C_s)$,~i.e. the textured mesh of $I_s$, and contains abundant yet compact spatial information (e.g., no background but adequate facial texture) to help estimate the desired face motion field.

\subsection{Dense Correspondence Estimator}\label{sec:dense_corr_esti}
Intuitively, the mismatch between $I_s$ and $I_d$ in both identity and image content (e.g., background and high-frequency texture)
makes it difficult to accurately estimate the desired face motion field for animating $I_s$.
However, according to the discussion in Section~\ref{sec:3d_face_mani_module}, the advantages of textured face proxies rendered by FLAME (i.e., $P_s$ and $P_r$) help mitigate those gaps and provide an aligned space for motion field estimation.
In this work, we model the face motion between $I_s$ and $I_d$ with a dense correspondence field $\mathcal{F}$, where each element $f_{i,j}\in\mathcal{F}$ denotes the correlation between two image patches $p_i$ and $p_j$ within $P_s$ and $P_r$, respectively.

Specifically, we use an encoder network $E_{corr}$ followed by two sequences of ResBlocks, denoted as $E_{res}^{(s)}$ and $E_{res}^{(r)}$, to extract features from $P_s$ and $P_r$, respectively.
Feature maps computed by a pre-trained HR-Net~\cite{sun2019deep} (denoted as $L_s$ and $L_r$) are also involved to represent the distribution of facial key-points, which help the network better explore the inner relationship between facial components.
So far, both face proxy ${P}$ and landmark embedding ${L}$ only describe the content within facial area, and do not contain the features of the rest of images.
Therefore, to make the network learn about the spatial layout beyond facial regions, we also feed the parsing map of $I_s$ (denoted as $M_s$) computed by a pre-trained BiSeNet~\cite{yu2018bisenet} to the feature extraction network (see Fig.~\ref{fig:SSS-GAN}).

Mathematically, the feature vector for the source face image could be written as
\begin{equation}
    f_s=E_{res}^{(s)}(E_{corr}(P_s), L_s, M_s)
\end{equation}
and similarly, we have
\begin{equation}
    f_r=E_{res}^{(r)}(E_{corr}(P_r), L_r)
\end{equation}
describing the feature of $P_r$.
Therefore, the dense correspondence field $\mathcal{F}$ could be represented by the correlation matrix between $f_s$ and $f_r$~\cite{zhang2020cross,zhou2021cocosnet}, which could be formulated as
\begin{equation}
    \label{eq:matrix_multiplication}
    \mathcal{F}=softmax(\hat{f}_r\cdot \hat{f}_s^T)\in \mathbb{R}^{hw\times hw}
\end{equation}
where $\hat{f}_s\in \mathbb{R}^{hw\times 1}$ and $\hat{f}_r\in \mathbb{R}^{hw\times 1}$ denote the flattened and normalized version of $f_s\in \mathbb{R}^{h\times w}$ and $f_r\in \mathbb{R}^{h\times w}$, respectively.

\begin{figure}[t]
    \centering
    \includegraphics[width=0.9\linewidth]{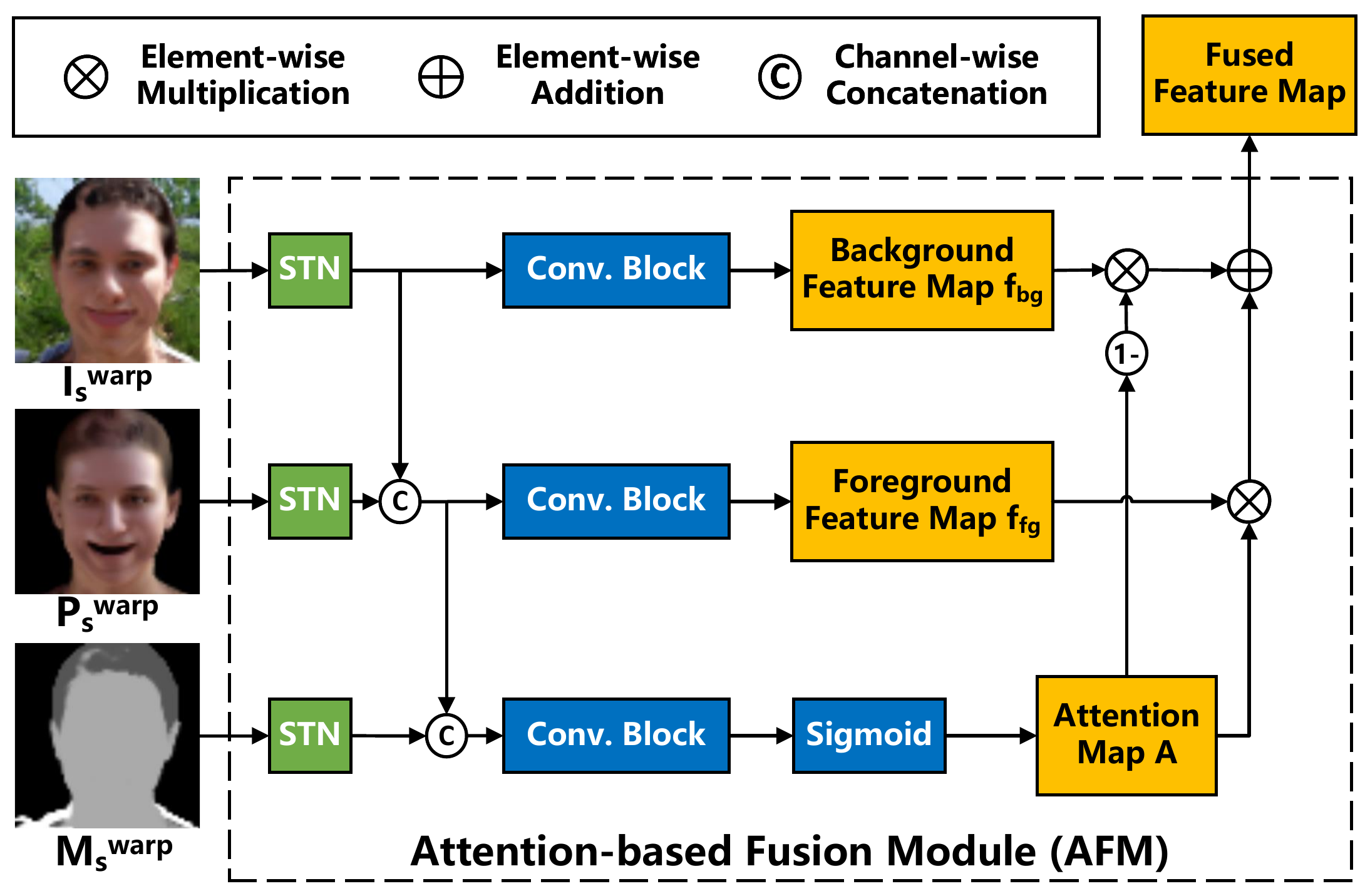}
    \caption{The network structure of AFM. Feature maps of foreground and background are fused with an attention map computed from all input, and the result is considered as the semantic guidance for the subsequent generation process.}
    \label{fig:AFM}
\end{figure}

Notably, with the column-wise softmax operation in Eq.~\ref{eq:matrix_multiplication}, warping of $I_s$ could simply be achieved by matrix multiplication,~i.e. $I_s^{warp}=\mathcal{F}\cdot\tilde{I}_s$, where $\tilde{I}_s\in\mathbb{R}^{hw\times1}$ denotes $I_s$ after down-sampling and flattening.
The advantage is two-fold: 1) it could be easily extended to other spatial representations of $I_s$,~e.g. textured face proxy $P_s$ ($P_s^{warp}=\mathcal{F}\cdot\tilde{P}_s$) or parsing map $M_s$ ($M_s^{warp}=\mathcal{F}\cdot\tilde{M}_s$), to provide multi-modal prior knowledge for the subsequent generation process (see Section~\ref{sec:facial_prior_guided_generator}), and 2) the reverse mapping could also be implemented with a simple matrix multiplication operation, which facilitates the computation of cycle consistency constraints (see Section~\ref{sec:objective_functions}).

\subsection{Facial Prior-guided Generator}\label{sec:facial_prior_guided_generator}
Although $I_s^{warp}$ could theoretically approximate the ideal re-enactment result and serve as a generative prior, it is often blurred due to the loss of high-frequency details in $I_s$, and distorted due to the error of estimated warping field $\mathcal{F}$ (see Fig.~\ref{fig:SSS-GAN}).
To solve this problem, we propose a Facial Prior-guided Generator to synthesize the final re-enactment result with high visual quality, which incorporates various warped spatial representations of the source subject (i.e., $I_s^{warp}$, $M_s^{warp}$, and $P_s^{warp}$) as the guiding prior knowledge.

As shown in Fig.~\ref{fig:SSS-GAN}, the generator takes the down-sampled concatenation of $I_s^{warp}$, $M_s^{warp}$, and $P_s^{warp}$ as input, and generates the final output with a sequence of ResBlocks.
For ResBlocks on coarse levels, $I_s^{warp}$, $M_s^{warp}$, and $P_s^{warp}$ are involved to introduce prior knowledge on the spatial layout of re-enacted faces.
Since they contain different semantic information of $I_r$, we design an Attention-based Fusion Module (AFM, as shown in Fig.~\ref{fig:AFM}) to fuse them into a single-channeled feature map, which is injected into the main body of the generator for semantic guidance.

Concretely, since the information of image background is only embedded in $I_s^{warp}$, while that of the foreground region (i.e., facial skin area) is contained in both $I_s^{warp}$ and $P_s^{warp}$, the feature of background ($f_{bg}$) and foreground ($f_{fg}$) could be computed by 
\begin{equation}
    f_{bg}=conv(STN(I_s^{warp}))
\end{equation}
\begin{equation}
    f_{fg}=conv(concat(STN(I_s^{warp}), STN(P_s^{warp})))
\end{equation}
where $STN$ denotes the Spatial Transformer Network~\cite{jaderberg2015spatial}, $conv$ stands for the convolutional block, and $concat$ refers to the channel-wise concatenation operation.
Then, $f_{bg}$ and $f_{fg}$ are merged by an attention map $A$ computed based on all three input, as shown in Fig.~\ref{fig:AFM}.
The final output $f_{out}$ could be obtained by merging $f_{bg}$ and $f_{fg}$ via
\begin{equation}
    f_{out}=A\cdot f_{fg} + (1-A)\cdot f_{bg}
\end{equation}

Afterwards, $f_{out}$ is integrated into the corresponding Resblock via spatially-adaptive normalization~\cite{park2019SPADE}.
In contrast, due to the inevitable error of warping field estimation, we do not incorporate $I_s^{warp}$, $M_s^{warp}$, and $P_s^{warp}$ into ResBlocks on finer levels, since they may increase the chance of introducing noise on high-frequency textural details.

\subsection{Objective Functions}\label{sec:objective_functions}
The dense correspondence estimator and facial prior-guided generator are trained simultaneously with the following objective functions.

\subsubsection{Losses on Dense Correspondence Estimation}
To regulate the estimated dense correspondence field $\mathcal{F}$, a straightforward pixel-level loss $L_{mw}$ is used to minimize the error of proxy warping, which could be written as
\begin{equation}
    L_{mw}=\Vert P_s^{warp} - P_r \Vert_1
\end{equation}
Similar to~\cite{zhu2017unpaired}, in order to regulate the mapping learned, a cycle consistency loss $L_{cc}$ is also adopted to ensure that the reconstructed source image $I_s'$ is close to $I_s$, which could be formulated as
\begin{equation}
    L_{cc}=\Vert I_s' - I_s \Vert_1
\end{equation}
$I_s'=softmax(\mathcal{F}^T)\cdot \tilde{I}^{warp}_s$ denotes the reverse mapping result of $I^{warp}_s$ obtained via matrix multiplication, which demonstrates the advantage of modeling the face motion with a correlation matrix.

\subsubsection{Losses on Semantic Consistency}
To fulfill the requirement of face re-enactment, explicit constraints $L_{id}$ and $L_{geo}$ are adopted to control the personal attributes and facial geometry of $I_r$, respectively.
Mathematically, the identity loss $L_{id}$ could be formulated as
\begin{equation}
    L_{id}=1-cos(\mathcal{A}(I_r), \mathcal{A}(I_s))
\end{equation}
where $\mathcal{A}$ is a pre-trained ArcFace~\cite{deng2019arcface} network and $cos(\cdot,\cdot)$ denotes the cosine similarity of two face embeddings. 
The geometry loss $L_{geo}$ could be written as
\begin{equation}
    L_{geo}=\Vert \beta_r - \beta_d \Vert_1 + \Vert \theta_r - \theta_d \Vert_1
\end{equation}
where $\beta_r$ ($\beta_d$) and $\theta_r$ ($\theta_d$) are 3DMM coefficients for pose and expression of $I_r$ ($I_d$) regressed by $E_{3D}$, respectively.

Unlike existing methods~\cite{ha2020marionette,wang2021one,yao2021one} merely using paired data $\langle I_s, I_s^\ast\rangle$ from the same video for image-level supervision, we adopt a hybrid learning scheme where $I_s$ and $I_d$ with different identity are also involved during the training process.
Therefore, the dense pixel-level loss $L_{pix}$ could be jointly formulated as
\begin{equation}
    L_{pix}= \left\{ \begin{array}{cl}
    \Vert I_r^\ast - I_s^\ast \Vert_2 + \Vert M^{warp}_s - \mathcal{M}(I_s^\ast) \Vert_2 & \mbox{if} \hspace{0.5em} I_d=I_s^\ast \\
                                    0 & \mbox{otherwise}
    \end{array}\right.
\end{equation}
where $\mathcal{M}$ denotes the pre-trained BiSeNet~\cite{yu2018bisenet} for computing the face parsing map.
The strategy for selecting $I_d$ will be detailed in Section~\ref{sec:setup}.



\subsubsection{Losses on Visual Fidelity}
Let us use $G$ to roughly refer to the entire network of DCG-GAN for face re-enactment,~i.e. $I_r=G(I_s, I_d)$.
To improve the visual quality of generation results, a discriminator network $D$ is adopted to distinguish synthetic images from realistic ones.
$D$ is designed to be of an ordinary network structure similar to PatchGAN~\cite{isola2017image}, and the hinge loss $h(t)=\min(0,-1+t)$ is used for discriminator regularization~\cite{wang2018high,zhang2020cross}.
Mathematically, the adversarial loss for $G$ and $D$ could be written as,
\begin{align}
\begin{split}
    L_{adv}^D = &-\mathbb{E}_{I_d\sim p_{data}}[h(D(I_d))] \\
                  &-\mathbb{E}_{I_s,I_d\sim p_{data}}[h(-D(G(I_s, I_d)))]
\end{split}\\
    L_{adv}^G = &-\mathbb{E}_{I_s,I_d\sim p_{data}}[D(G(I_s, I_d))]
\end{align}
where $p_{data}$ refers to the real data distribution.

\subsubsection{The Overall Training Objective}
In summary, the overall training objective function for $G$ and $D$ could be written as
\begin{align}
    \begin{split}
        L_G = \hspace{0.3em} & \lambda_{mw}L_{mw} + \lambda_{cc}L_{cc} + \\
              & \lambda_{id}L_{id} + \lambda_{geo}L_{geo} + \lambda_{pix}L_{pix} + \\
              & \lambda_{adv_G} L_{adv}^G
    \end{split} \\
        L_D = \hspace{0.3em} & L_{adv}^D
\end{align}
where $\lambda_{mw}$, $\lambda_{cc}$, $\lambda_{id}$, $\lambda_{geo}$, $\lambda_{pix}$, and $\lambda_{adv}$ are coefficients balancing the important of different loss terms. $L_G$ and $L_D$ are minimized alternatively until convergence.

\begin{figure*}[t]
    \centering
    \includegraphics[width=0.85\linewidth]{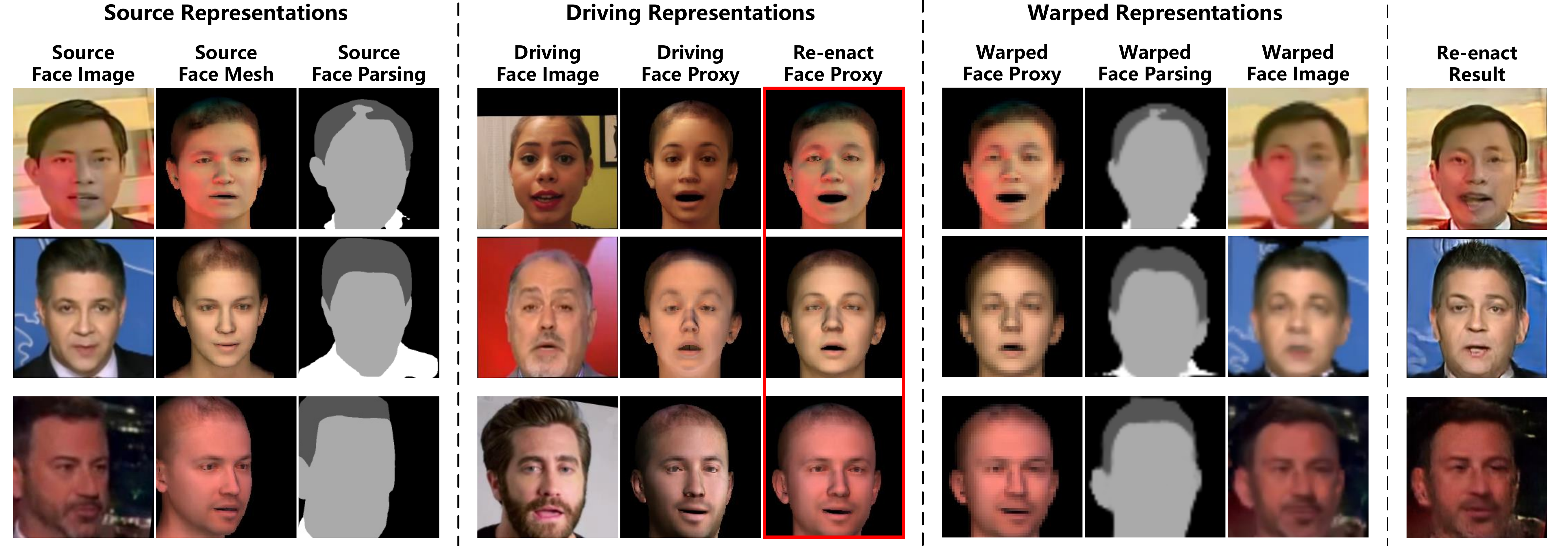}
    \caption{Visualization of intermediate warping results. It is clear that the warped representations successfully predict the semantic of re-enacted faces.}
    \label{fig:quanlitative_results}
\end{figure*}

\section{Experiments}\label{sec:experiments}
\subsection{Configuration Setup}\label{sec:setup}
\noindent\textbf{Implementation Details}
The proposed model is trained by an Adam optimizer with $\beta_1=0.9$, $\beta_2=0.999$, and the learning rate is set to $1e-4$.
We empirically set $\lambda_{geo}$ to $1.0$, and roughly adjust other coefficients so that the corresponding loss value is one order of magnitude lower.
At each iteration, the network is trained with paired data from the same video by a probability of $p\in[0,1]$, and with unpaired face images by the chance of $1-p$.
To facilitate convergence, $p$ is set to a high value (e.g., 0.8) in starting epochs, and gradually drops to accustom the network to unpaired input.
A NVIDIA A100 GPU is used to run all experiments with the batch size set to 8.

\noindent\textbf{Datasets}
Experiments are conducted on the following datasets with all images resized to $256\times 256$.
\begin{itemize}
    
    
    \item \textbf{FF++}~\cite{rossler2019faceforensics++}: FaceForensics++ (FF++) contains 1,000 video sequences of near frontal face without occlusion. We evenly extract 10 frames from each video and collect 10,000 images in total.
    
    \item \textbf{Celeb-DF}~\cite{li2020celeb}: Celeb-DF is a large-scale dataset containing 890 realistic videos collected from YouTube, which is challenging as it covers a wide range of age, gender, and ethnic groups with extreme poses and varying image quality. 
    
    \item \textbf{VoxCeleb1}~\cite{nagrani2017voxceleb}: VoxCeleb1 is another large-scale video dataset of 1.251 celebrities acquired from YouTube. We directly use the cropped images released on the official website~\footnote{https://www.robots.ox.ac.uk/$\sim$vgg/data/voxceleb/vox1.html} in our experiments.
\end{itemize}

Similar to~\cite{siarohin2019first}, on FF++ and Celeb-DF, we randomly select videos of $80\%$ subjects for training, and use the rest data for testing with no identity overlap with the training split. 
VoxCeleb1 is used as a driving dataset to test the performance of cross-domain face re-enactment where source images are selected from the testing split of FF++ and Celeb-DF.

\noindent\textbf{Benchmark Methods}
In order to achieve fair performance comparison, we choose four representative methods with code released as the benchmark, including One-shot~\cite{OneShotFace2019}, Latent-pose~\cite{burkov2020neural}, Bi-layer~\cite{zakharov2020fast}, and FOMM~\cite{siarohin2019first}.
Note that there are also other approaches proposed for one-shot face re-enactment (e.g.,~\cite{yao2021one,yao2020mesh}). However,
no publicly available code has been found for these methods, and it would be unfair if we compare to our re-implementations.

\noindent\textbf{Evaluation Metrics}
As for quantitative evaluation, three metrics commonly used in previous one-shot face re-enactment studies~\cite{ha2020marionette,wang2021one,yao2021one,OneShotFace2019} are adopted for performance evaluation.
Cosine similarity (\textbf{CosSim}) between the feature of $I_r$ and $I_s$ are computed to evaluate the performance on identity preservation.
Mean Squared Error (MSE) of the pose (\textbf{PoseMSE}) and expression (\textbf{ExpMSE}) embedding between $I_r$ and $I_d$ measures the accuracy of re-enactment fulfillment.
We use CosFace~\cite{wang2018cosface} and Hopenet~\cite{ruiz2018fine} to extract the feature for identity and pose, respectively.
To provide a more comprehensive evaluation, both 3DDFA~\cite{zhu2017face} and FaceWarehouse~\cite{cao2013facewarehouse} are adopted to compute the embedding vector for expression.

\subsection{Qualitative Results}
\noindent\textbf{Single-domain face re-enactment.} Sample results of one-shot face re-enactment on FF++ and Celeb-DF are shown in Fig.~\ref{fig:quanlitative_results}.
It is clear that the textured proxy of re-enacted faces (i.e., $P_r$, in the red box) are able to convey the pose and expression of driving faces, while maintain the identity and lighting condition in source images.
The key to this success is that 3DMM provides us with a set of disentangled geometric basis, which serves as a common coordinate system where we could project face images and obtain aligned coefficient representations.
As a result, it improves the accuracy of dense correspondence estimation by eliminating the interference of both driving identity and redundant high-frequency details in non-facial area.
Warped representations shown in Fig.~\ref{fig:quanlitative_results} indicate that the computed correspondence field $\mathcal{F}$ is also effective beyond the facial region, thanks to the prior knowledge (i.e., $M_s$, $L_s$, and $L_r$) introduced when extracting features for $P_s$ and $P_r$.

\begin{figure*}[t]
    \centering
    \includegraphics[width=0.95\linewidth]{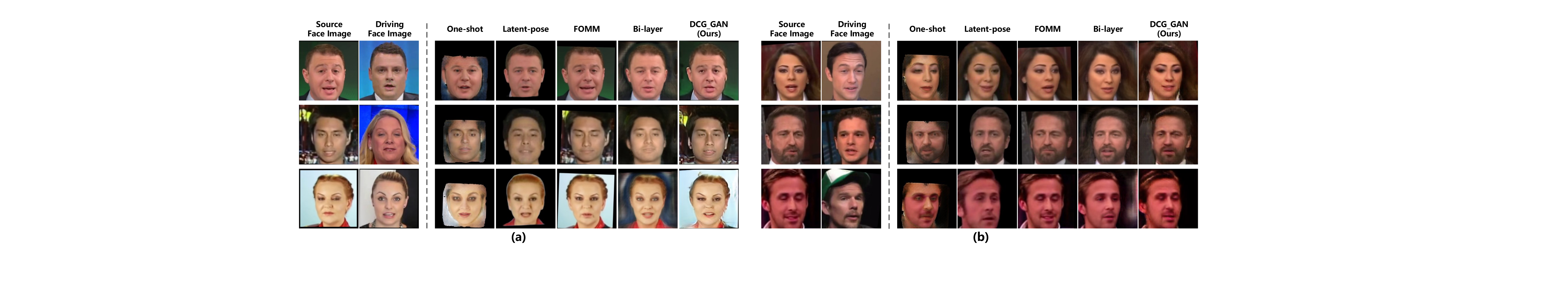}
    \caption{Sample results of single-domain face re-enactment on FF++ and Celeb-DF. All images are aligned according to eye positions for ease of comparison.}
    \label{fig:single_dataset_comparison}
\end{figure*}

\begin{figure*}[t]
    \centering
    \includegraphics[width=0.95\linewidth]{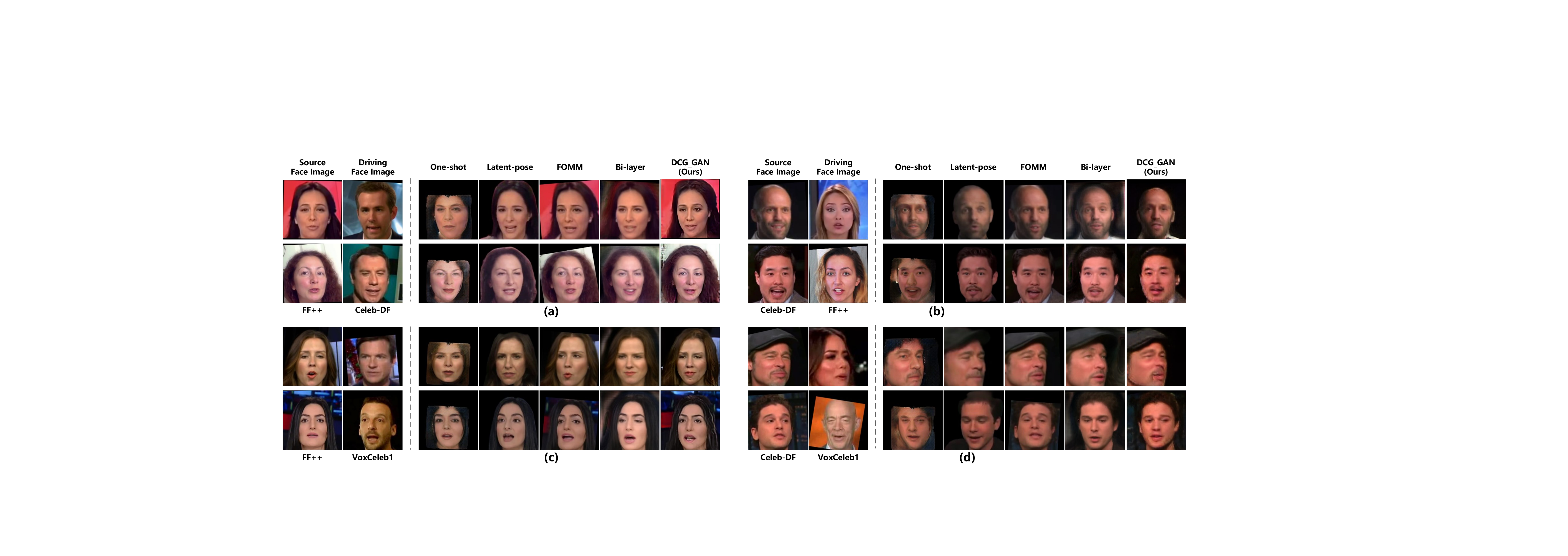}
    \caption{Sample results of cross-domain face re-enactment with the name of dataset labeled underneath. All images are aligned according to eye positions for ease of comparison.}
    \label{fig:cross_dataset_comparison}
\end{figure*}

According to comparison results shown in Fig.~\ref{fig:single_dataset_comparison}, both One-shot and Latent-pose fail to completely disentangle the identity information in driving faces, and thus they interfere with the generation results and cause large face shape distortion.
Although FOMM shows better performance in identity preservation, it fails to accurately capture pose changes and subtle movement of facial components (especially for eyes and mouth).
Also note that FOMM requires image sequences as input for driving at inference time, while our method could work in cases where only one single driving image is available.
Moreover, our method outperforms Bi-layer in synthesizing more realist hair region and background texture preservation.

\noindent\textbf{Cross-domain face re-enactment.}
We also test the performance of DCG-GAN when driving images are sampled from unseen dataset, which is very common in practical applications.
According to results shown in Fig.~\ref{fig:cross_dataset_comparison}, our method could well-adapt to the discrepancy of data distributions.
This is because 3DMM could also help eliminate the interference of domain-specific data bias, besides the identity of driving subjects.
Inaccurately manipulated pose and expression could be clearly observed in results obtained by One-shot and FOMM.
Although Latent-pose achieves more satisfying performance in re-enactment fulfillment, it still suffers from large identity shift.
Similar to results in single-domain face re-enactment, Bi-layer fails to handle the translation of non-facial region.

\begin{table*}[t]
\caption{Quantitative results of both single- and cross-domain face re-enactment. Elements in angled brackets denote the name of datasets where source (first) and driving (second) are sampled. ExpMSE\textsubscript{1} and ExpMSE\textsubscript{2} denote the MSE between expression embeddings compute by 3DFFA and Facewarehouse, respectively. Scores ranked in the first and second places are highlighted in {\color{red}\textbf{red}} and {\color{blue}\textbf{blue}} under each metric (Note: One-shot\textsuperscript{*} and Latent-pose\textsuperscript{*} are not considered due to their poor performance in identity preservation).}
\label{table:quantitative_results}
\centering
\resizebox{1.95\columnwidth}{!}{%
\ra{1.1}
\begin{tabular} {@{}l llll c llll@{}}
\toprule
 & \multicolumn{1}{c}{CosSim$\uparrow$} & \multicolumn{1}{c}{PoseMSE$\downarrow$} & \multicolumn{1}{c}{ExpMSE\textsubscript{1}$\downarrow$} & \multicolumn{1}{c}{ExpMSE\textsubscript{2}$\downarrow$} & \phantom{a} 
 & \multicolumn{1}{c}{CosSim$\uparrow$} & \multicolumn{1}{c}{PoseMSE$\downarrow$} & \multicolumn{1}{c}{ExpMSE\textsubscript{1}$\downarrow$} & \multicolumn{1}{c}{ExpMSE\textsubscript{2}$\downarrow$} \\
\midrule
 & \multicolumn{4}{c}{\textlangle\hspace{0.2em} FF++, FF++ \textrangle} &\phantom{a} & \multicolumn{4}{c}{\textlangle\hspace{0.2em} Celeb-DF, Celeb-DF \textrangle} \\
   \cmidrule{2-5}                          \cmidrule{7-10}  
One-shot\textsuperscript{*}    & $0.34\pm 0.12$ & $3.63\pm 1.88$  & $0.98\pm 0.41$ & $3.41\pm 0.68$ &  & $0.29\pm 0.12$ & $5.24\pm 2.88$  & $1.14\pm 0.44$ & $3.41\pm 0.68$ \\
Latent-pose\textsuperscript{*} & $0.37\pm 0.12$ & $10.25\pm 5.87$ & $1.33\pm 0.61$ & $3.48\pm 0.64$ &  & $0.36\pm 0.12$ & $12.13\pm 8.67$ & $1.33\pm 0.62$ & $3.20\pm 0.60$ \\
Bi-layer & $0.61\pm 0.13$ & $9.14\pm 3.76$ & $\mathbf{\textcolor{blue}{1.26}}\pm 0.56$ & $\mathbf{\textcolor{red}{2.93}}\pm 0.58$ &  & $0.56\pm 0.15$ & $11.73\pm 5.83$ & $1.26\pm 0.53$ & $\mathbf{\textcolor{red}{3.03}}\pm 0.58$ \\
FOMM        & $\mathbf{\textcolor{red}{0.91}}\pm 0.08$ & $11.42\pm 6.55$ & $1.40\pm 0.66$ & $3.61\pm 0.69$ &  & $\mathbf{\textcolor{red}{0.83}}\pm 0.12$ & $\mathbf{\textcolor{blue}{11.89}}\pm 8.60$ & $1.40\pm 0.60$ & $\mathbf{\textcolor{blue}{3.53}}\pm 0.65$ \\
DCG-GAN-sD  & $0.68\pm 0.11$ & $\mathbf{\textcolor{blue}{8.82}}\pm 4.94$  & $1.24\pm 0.55$ & $3.48\pm 0.70$ &  & $0.50\pm 0.14$ & $14.92\pm 5.92$ & $1.42\pm 0.66$ & $3.61\pm 0.69$ \\
DCG-GAN-sG  & $0.22\pm 0.11$ & $8.86\pm 4.55$  & $\mathbf{\textcolor{blue}{1.22}}\pm 0.52$ & $3.46\pm 0.67$ &  & $0.06\pm 0.08$ & $12.49\pm 5.58$ & $1.49\pm 0.66$ & $3.84\pm 0.66$ \\
DCG-GAN-sDG & $0.07\pm 0.09$ & $9.88\pm 5.13$  & $1.25\pm 0.58$ & $3.48\pm 0.72$ &  & $0.06\pm 0.09$ & $15.70\pm 5.96$ & $1.56\pm 0.60$ & $3.95\pm 0.75$ \\
DCG-GAN & $\mathbf{\textcolor{blue}{0.84}}\pm 0.12$ & $\mathbf{\textcolor{red}{8.50}}\pm 4.52$  & $\mathbf{\textcolor{red}{1.21}}\pm 0.51$ & $\mathbf{\textcolor{blue}{3.42}}\pm 0.70$ &  & $\mathbf{\textcolor{blue}{0.79}}\pm 0.17$ & $\mathbf{\textcolor{red}{10.92}}\pm 5.20$ & $\mathbf{\textcolor{red}{1.25}}\pm 0.58$ & $3.55\pm 0.68$ \\
&  &  &  &  &  &  &  &  & \\[-2ex]
 & \multicolumn{4}{c}{\textlangle\hspace{0.2em} FF++, Celeb-DF \textrangle} &\phantom{a} & \multicolumn{4}{c}{\textlangle\hspace{0.2em} Celeb-DF, FF++ \textrangle} \\
   \cmidrule{2-5}                          \cmidrule{7-10}  
One-shot\textsuperscript{*} & $0.31\pm 0.12$ & $4.70\pm 2.76$  & $1.09\pm 0.42$ & $3.48\pm 0.73$ &  & $0.31\pm 0.13$ & $4.40\pm 2.29$  & $1.10\pm 0.43$ & $3.34\pm 0.66$ \\
Latent-pose\textsuperscript{*} & $0.37\pm 0.12$ & $12.37\pm 8.62$ & $1.35\pm 0.66$ & $3.26\pm 0.65$ &  & $0.36\pm 0.12$ & $11.22\pm 6.47$ & $1.37\pm 0.65$ & $3.20\pm 0.66$ \\
Bi-layer & $0.60\pm 0.13$ & $10.89\pm 6.25$ & $\mathbf{\textcolor{blue}{1.40}}\pm 0.59$ & $\mathbf{\textcolor{blue}{3.46}}\pm 0.64$ &  & $0.59\pm 0.14$ & $\mathbf{\textcolor{blue}{11.01}}\pm 4.35$ & $\mathbf{\textcolor{blue}{1.36}}\pm 0.56$ & $\mathbf{\textcolor{blue}{3.57}}\pm 0.62$ \\
FOMM     & $\mathbf{\textcolor{red}{0.84}}\pm 0.11$ & $\mathbf{\textcolor{blue}{10.30}}\pm 5.61$ & $1.43\pm 0.66$ & $3.59\pm 0.74$ &  & $\mathbf{\textcolor{red}{0.90}}\pm 0.09$ & $14.66\pm 8.54$ & $1.50\pm 0.70$ & $3.65\pm 0.71$ \\
DCG-GAN & $\mathbf{\textcolor{blue}{0.77}}\pm 0.17$ & $\mathbf{\textcolor{red}{10.01}}\pm 5.54$ & $\mathbf{\textcolor{red}{1.36}}\pm 0.60$ & $\mathbf{\textcolor{red}{3.43}}\pm 0.71$ &  & $\mathbf{\textcolor{blue}{0.82}}\pm 0.17$ & $\mathbf{\textcolor{red}{10.75}}\pm 6.05$ & $\mathbf{\textcolor{red}{1.31}}\pm 0.58$ & $\mathbf{\textcolor{red}{3.54}}\pm 0.66$ \\
 &  &  &  &  &  &  &  &  & \\[-2ex]
 & \multicolumn{4}{c}{\textlangle\hspace{0.2em} FF++, VoxCeleb1 \textrangle} &\phantom{a} & \multicolumn{4}{c}{\textlangle\hspace{0.2em} Celeb-DF, VoxCeleb1 \textrangle} \\
   \cmidrule{2-5}                          \cmidrule{7-10}  
One-shot\textsuperscript{*} & $0.30\pm 0.12$ & $5.00\pm 2.84$  & $1.09\pm 0.44$ & $3.48\pm 0.73$ &  & $0.28\pm 0.13$ & $5.35\pm 2.97$  & $1.11\pm 0.44$ & $3.45\pm 0.69$ \\
Latent-pose\textsuperscript{*} & $0.36\pm 0.12$ & $12.14\pm 7.67$ & $1.36\pm 0.66$ & $3.27\pm 0.68$ &  & $0.36\pm 0.12$ & $12.98\pm 8.68$ & $1.36\pm 0.64$ & $3.22\pm 0.70$ \\
Bi-layer & $0.59\pm 0.13$ & $11.46\pm 6.32$ & $\mathbf{\textcolor{blue}{1.33}}\pm 0.58$ & $3.61\pm 0.64$ &  & $0.58\pm 0.15$ & $11.19\pm 6.09$ & $\mathbf{\textcolor{blue}{1.37}}\pm 0.56$ & $\mathbf{\textcolor{blue}{3.47}}\pm 0.67$ \\
FOMM     & $\mathbf{\textcolor{blue}{0.80}}\pm 0.13$ & $\mathbf{\textcolor{blue}{10.35}}\pm 5.81$ & $1.43\pm 0.66$ & $\mathbf{\textcolor{blue}{3.59}}\pm 0.74$ &  & $\mathbf{\textcolor{red}{0.79}}\pm 0.14$ & $\mathbf{\textcolor{blue}{10.18}}\pm 7.16$ & $1.38\pm 0.65$ & $3.50\pm 0.71$ \\
DCG-GAN & $\mathbf{\textcolor{red}{0.81}}\pm 0.11$ & $\mathbf{\textcolor{red}{10.15}}\pm 5.41$ & $\mathbf{\textcolor{red}{1.31}}\pm 0.58$ & $\mathbf{\textcolor{red}{3.55}}\pm 0.70$ &  & $\mathbf{\textcolor{blue}{0.74}}\pm 0.17$ & $\mathbf{\textcolor{red}{9.81}}\pm 5.20$  & $\mathbf{\textcolor{red}{1.31}}\pm 0.60$ & $\mathbf{\textcolor{red}{3.42}}\pm 0.75$ \\
\bottomrule
\end{tabular}
}
\end{table*}

\subsection{Quantitative Results}\label{sec:quantitative_results}
Quantitative results for both single- and cross-domain face re-enactment are reported in Table~\ref{table:quantitative_results}.
Since One-shot enforces the shape of re-enacted faces close to that of driving images, it produces lower error in re-enactment fulfillment but completely fails to preserve the identity of source faces (CosSim values lower than 0.35 in all cases).
Latent-pose also suffers from poor identity preservation performance, where CosSim values are around 0.36 to 0.37.
In contrast, our method achieves much better performance in preserving the source identity, indicating the success of using 3DMM to bridge the identity gap in one-shot face re-enactment.
Although CosSim scores obtained by FOMM are slightly higher than ours, it produces larger errors in pose and expression transformation compared to our methods in most cases.
Moreover, FOMM requires a sequence of faces with no abrupt change for driving, while our method only needs one driving image to achieve face re-enactment, which has a much broader range of application.

\begin{figure}[t]
    \centering
    \includegraphics[width=0.90\linewidth]{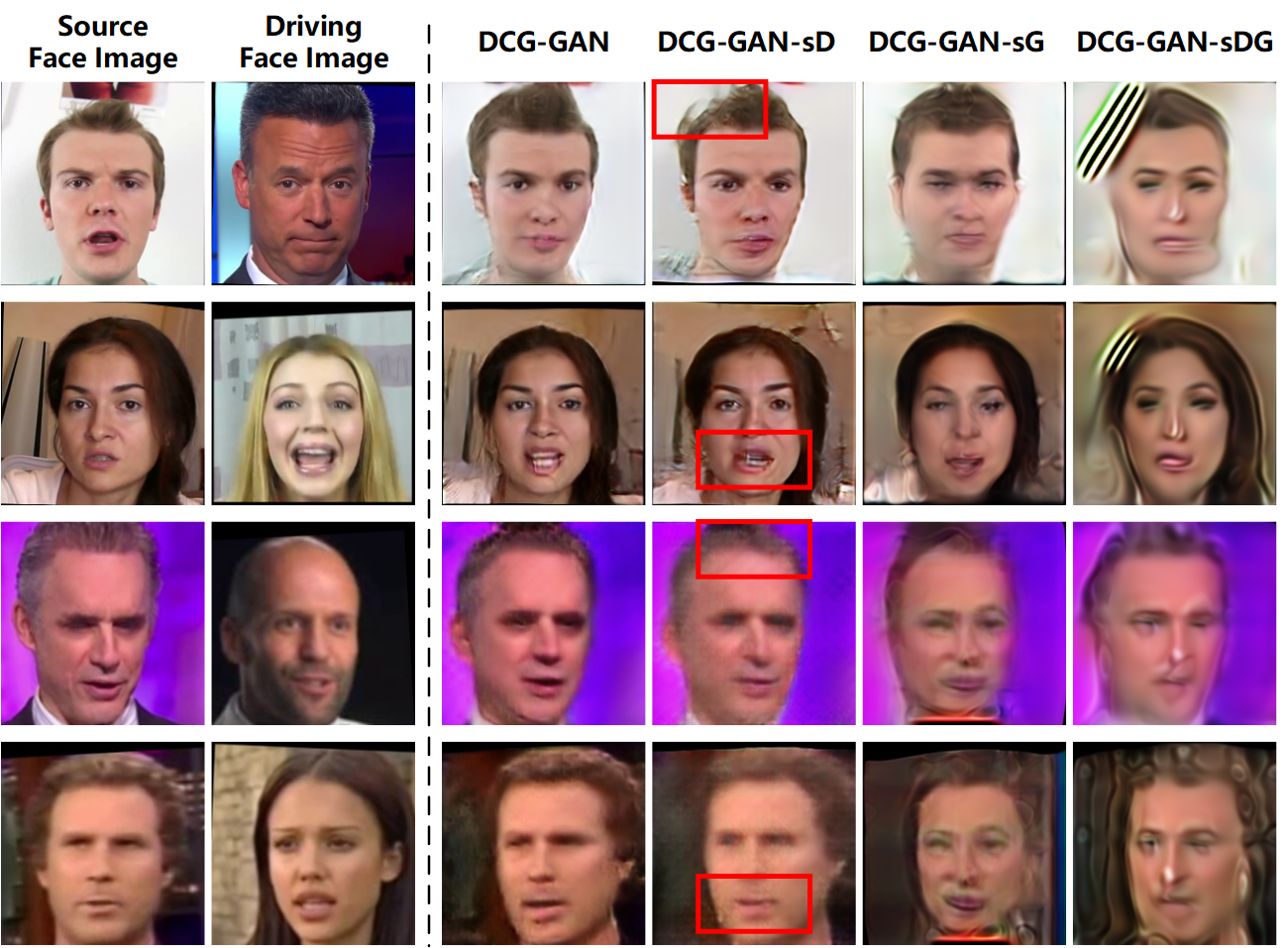}
    \caption{Sample results obtained by DCG-GAN and its variants on (a) FF++ and (b) Celeb-DF. Red boxes encircle image distortion caused by using the simplified dense correspondence estimation module.}
    \label{fig:ablation_study}
\end{figure}

\subsection{Ablation Study}\label{sec:ablation_study}
In this subsection, we validate the effectiveness of the network design for each component in our method.
Specifically, we propose the following variants of DCG-GAN for investigation:
1) In \textbf{DCG-GAN-sD} (simple Dense correspondence estimation), $\mathcal{F}$ is computed merely based on $M_s$ and $M_r$ with no $L_s$, $L_r$ or $P_s$ involved;
2) \textbf{DCG-GAN-sG} (simple Generator) refers to the setup where no AFM network is used to guide the generation process, and $I_s^{warp}$, $M_s^{warp}$, $P_s^{warp}$ are fed to the generator only at the bottom layer;
and 3) \textbf{DCG-GAN-sDG} combines the configuration in both previous settings for comparison.

As shown in Fig.~\ref{fig:ablation_study}, removing the facial landmark ($L_s$ and $L_r$) and parsing map ($P_s$) impairs the accuracy of motion field estimation, especially for hair and mouth regions. 
The reason for this might be that the model fails to extract sufficient information of both the image layout (contained in $P_s$) and shape of facial components (embedded in $L_s$ and $L_r$).
Results obtained by DCG-GAN-sG have severe problem in identity preservation, as no detailed textural information is provided for ResBlocks at higher resolutions.
DCG-GAN-sDG is very unstable to train and we only present results obtained by the latest checkpoint before collapse, which suffer from severe artifacts.
Quantitative results of reported in Table~\ref{table:quantitative_results} are in line with our previous observations.
Our model consistently outperforms its variants under all measurements, indicating the effectiveness of the network design for each module.

\section{Conclusion}\label{sec:conclusion}
In this paper, we propose a novel GAN-based model, named DCG-GAN, for one-shot face re-enactment.
DCG-GAN adopts a 3DMM model to decompose and reorganize geometric semantics of faces, and generate textured mesh images to help estimate the dense correspondence between source and re-enacted faces.
Such dense correspondence field could be used to warp source images for re-enactment results approximation, followed by a generator network to improve the visual quality.
Extensive experiments on multiple datasets demonstrate the effectiveness of DCG-GAN in generating lifelike images with accurate re-enactment fulfillment.
We also discuss limitations of DCG-GAN and propose possible solutions as future research directions.

\bibliography{aaai23}

\end{document}